\pdfoutput=1

\documentclass[11pt]{article}

\usepackage{acl}

\usepackage{times}
\usepackage{latexsym}

\usepackage[T1]{fontenc}

\usepackage[utf8]{inputenc}

\usepackage{microtype}

\usepackage{inconsolata}

\usepackage{caption}
\usepackage{subcaption}
\usepackage{graphicx}

\usepackage{multirow,tabularx}

\title{A Comparison of Independent and Joint Fine-tuning Strategies for Retrieval-Augmented Generation}

\author{
    \textbf{Neal Lawton},
    \textbf{Alfy Samuel},
    \textbf{Anoop Kumar},
    \textbf{Daben Liu}
    \\
    \{neal.lawton, alfy.samuel, anoop.kumar, daben.liu\}@capitalone.com
}

\begin{document}
\maketitle
\begin{abstract}
Retrieval augmented generation (RAG) is a popular framework for question answering that is powered by two large language models (LLMs): an embedding model that retrieves context documents from a database that are relevant to a given question, and a generator model that uses the retrieved context to generate an answer to the question. Both the embedding and generator models can be fine-tuned to increase performance of a RAG pipeline on a new task, but multiple fine-tuning strategies exist with different costs and benefits. In this paper, we evaluate and compare several RAG fine-tuning strategies, including independent, joint, and two-phase fine-tuning. In our experiments, we observe that all of these strategies achieve about equal improvement in EM and F1 generation quality metrics, although they have significantly different computational costs. We conclude the optimal fine-tuning strategy to use depends on whether the training dataset includes context labels and whether a grid search over the learning rates for the embedding and generator models is required.
\end{abstract}

\section{Introduction}
\begin{figure*}[t]
    \centering
    \begin{subfigure}[b]{0.475\textwidth}
        \centering
        \includegraphics[width=\textwidth]{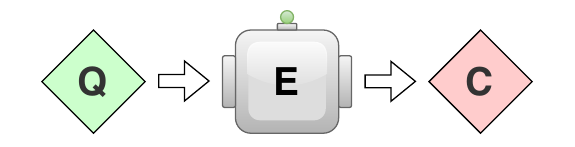}
        \caption{Fine-tune the embedding model using context labels.}%
        \label{fig:ft_embed}
    \end{subfigure}
    \hfill
    \begin{subfigure}[b]{0.475\textwidth}  
        \centering         
        \includegraphics[width=\textwidth]{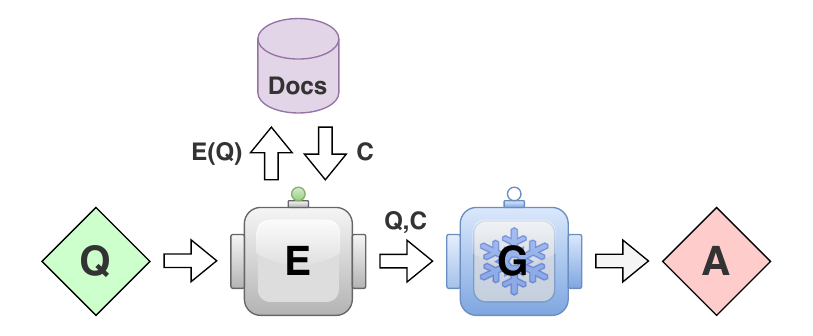}
        \caption{Freeze the generator model while fine-tuning the embedding model with either RAG-Token or RAG-Sequence.}%
        \label{fig:freeze_gen}
    \end{subfigure}
    \vskip\baselineskip
    \begin{subfigure}[b]{0.475\textwidth}   
        \centering 
        \includegraphics[width=\textwidth]{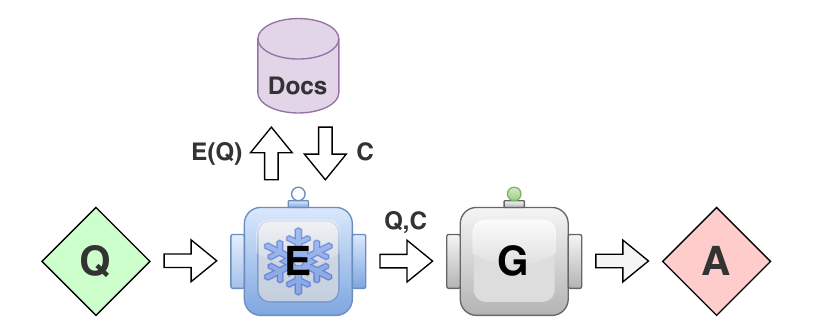}
        \caption{Freeze the embedding model while fine-tuning the generator model with RAG-Token or RAG-Sequence.}%
        \label{fig:freeze_embed}
    \end{subfigure}
    \hfill
    \begin{subfigure}[b]{0.475\textwidth}   
        \centering 
        \includegraphics[width=\textwidth]{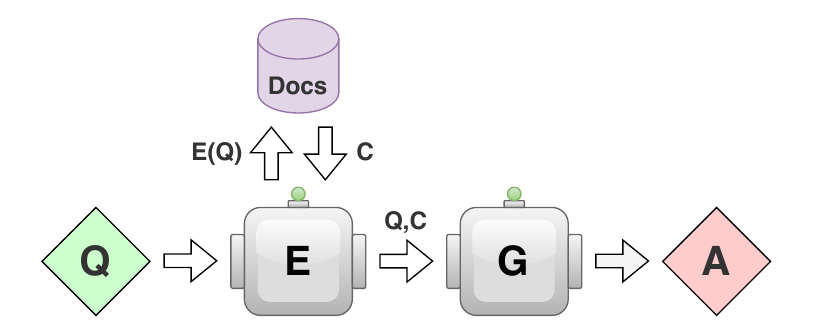}
        \caption{Fine-tune the embedding and generator models jointly with RAG-Token or RAG-Sequence.}%
        \label{fig:joint_ft}
    \end{subfigure}
    \caption{RAG fine-tuning strategy subprocesses. Each of the RAG fine-tuning strategies discussed in this paper uses a combination of these subprocesses. Key: \textbf{Q}uestion, \textbf{C}ontext, \textbf{A}nswer, \textbf{E}mbedding model, \textbf{G}enerator model.}
    \label{fig:rag_illustration}
\end{figure*}
Retrieval augmented generation (RAG) is a popular framework for  NLP tasks like question answering. RAG is powered by two LLMs: an embedding model that retrieves context documents from a database that are relevant to a given question, and a generator model that uses the retrieved context documents to generate an answer to the question. 

Both the embedding model and generator model can be fine-tuned to improve the end-to-end performance of a RAG pipeline. Given a dataset of (\textit{question}, \textit{context}) pairs, the embedding model can be fine-tuned to retrieve more relevant context documents for a given question. This requires a training dataset with context labels, i.e., where each question is paired with one or more relevant context documents from the database. Given a dataset of (\textit{question}, \textit{context}, \textit{answer}) triplets, where the context is either provided as part of the training dataset as context labels or retrieved from the database using a baseline embedding model, the generator model can be fine-tuned to increase the likelihood of generating the correct answer given the question and relevant context documents.

Although the embedding and generator models can be fine-tuned independently, fine-tuning both models jointly with an end-to-end fine-tuning method such as RAG-Token or RAG-Sequence \citep{lewis2020retrieval} may yield equal or better end-to-end performance without the need for context labels. Additionally, we consider a two-phase fine-tuning strategy that uses RAG-Token to first fine-tune the generator model while holding the embedding model frozen, then fine-tunes the embedding model while holding the generator model frozen. 

The choice of learning rate used for fine-tuning may significantly affect the end-to-end performance of the RAG pipeline, and the optimal choice of learning rate for the embedding and generator models may be different. We use a grid search to find a suitable choice of learning rates.

In this paper, we compare independent, joint, and two-phase fine-tuning and find they all achieve similar end-to-end performance when using a suitable choice of learning rates. Based on our experimental results, we make the following conclusions:
\begin{itemize}
\item 
Independent fine-tuning is the least computationally expensive strategy, and so should be used when possible. However, this strategy can only be used if the training dataset includes context labels.
\item
If context labels are not available, but a suitable choice of learning rate for the embedding and generator models is already known, then joint fine-tuning should be used since it is less computationally expensive than two-phase fine-tuning.
\item If context labels are not available and a suitable choice of learning rates for the embedding and generator models is unknown, then two-phase fine-tuning should be used while performing independent grid searches over the learning rates for the embedding and generator models. 
\end{itemize}

\section{Fine-tuning Strategies}
\subsection{Embedding Model Fine-tuning}
The embedding model of a RAG pipeline can be fine-tuned to retrieve more relevant context documents given a dataset of (\textit{question}, \textit{context}) pairs by minimizing the distance (or maximizing the similarity) between the embedding vectors of each (\textit{question}, \textit{context}) pair. This method is illustrated in Figure \ref{fig:ft_embed}. Note that the embedding vectors of the context documents are held frozen in the precomputed vector database, so that only the embedding vectors of the questions are updated. There are many different options for the choice of loss function to minimize, including contrastive loss \citep{hadsell2006dimensionality}, multiple negatives ranking loss \citep{henderson2017efficient}, and the GISTEmbed loss \citep{solatorio2024gistembed} using either cosine similarity or $L_2$ distance as the distance metric. Cached variants \citep{gao2021scaling} of these methods exist that allow for effectively much larger batch sizes without increased GPU memory usage. In our experiments, we use cosine similarity as the distance metric and multiple negatives ranking loss without caching with batch size 8 as the loss function.

\subsection{Generator Model Fine-tuning}
The generator model can be fine-tuned by minimizing the negative log-likelihood of the answer given the question and relevant context documents. In our experiments, we always fine-tune the generator model using context retrieved by a baseline embedding model rather than context labels. This is equivalent to the "frozen embedding" fine-tuning process illustrated in Figure \ref{fig:freeze_embed}. In our experiments, we fine-tune the generator model with QLoRA \citep{dettmers2023qlora, hu2022lora} using LoRA rank 16 and 4-bit quantization. 

\subsection{Joint Fine-tuning}
The embedding and generator models can be fine-tuned jointly by fine-tuning the RAG pipeline end-to-end with either RAG-Token or RAG-Sequence \citep{lewis2020retrieval}, illustrated in Figure \ref{fig:joint_ft}. Both these methods optimize an objective that is fully differentiable with respect to both the embedding model and generator model's parameters by approximating the RAG pipeline with a simplified probability model; the two methods differ only in the approximation they make. Instead of using context labels, these methods use context retrieved by the embedding model to fine-tune the generator model, and reward the embedding model for retrieving context documents that actually improve the generator model's prediction for the answer. In our experiments, we use full fine-tuning for the embedding model and QLoRA for the generator model. We fine-tune using two learning rates: one for the embedding model's parameters, and the other for the generator model's parameters.

\subsection{Two-Phase Fine-tuning}
We also consider a two-phase fine-tuning strategy that uses RAG-Token to first fine-tune the generator model while holding the embedding model frozen as in Figure \ref{fig:freeze_embed}, then fine-tunes the embedding model while holding the generator model frozen as in Figure \ref{fig:freeze_gen}. As in joint fine-tuning, we fine-tune using two learning rates.

\subsection{Learning Rate Grid Search}
Using a suitable choice of learning rate is important for maximizing end-to-end performance for each fine-tuning strategy. In order to find a near-optimal choice of learning rate, we perform a grid search over the learning rate for each experiment. Performing this grid search is computationally inexpensive for strategies that fine-tune only either the embedding model or generator model: we simply repeat the experiment for each grid value, then keep only the result that achieves the best end-to-end validation performance. The grid search is also computationally inexpensive when fine-tuning both models independently or with the two-phase strategy, since the grid search can be performed independently for the embedding and generator models. However, jointly optimizing over the learning rates for the embedding and generator models is much more computationally expensive. Instead, in our joint fine-tuning experiments, we use the same learning rates as those discovered by the grid search for the two-phase fine-tuning strategy.

\section{Experiments}
\begin{figure*}[t]
  \centering
  \includegraphics[width=0.9\linewidth]{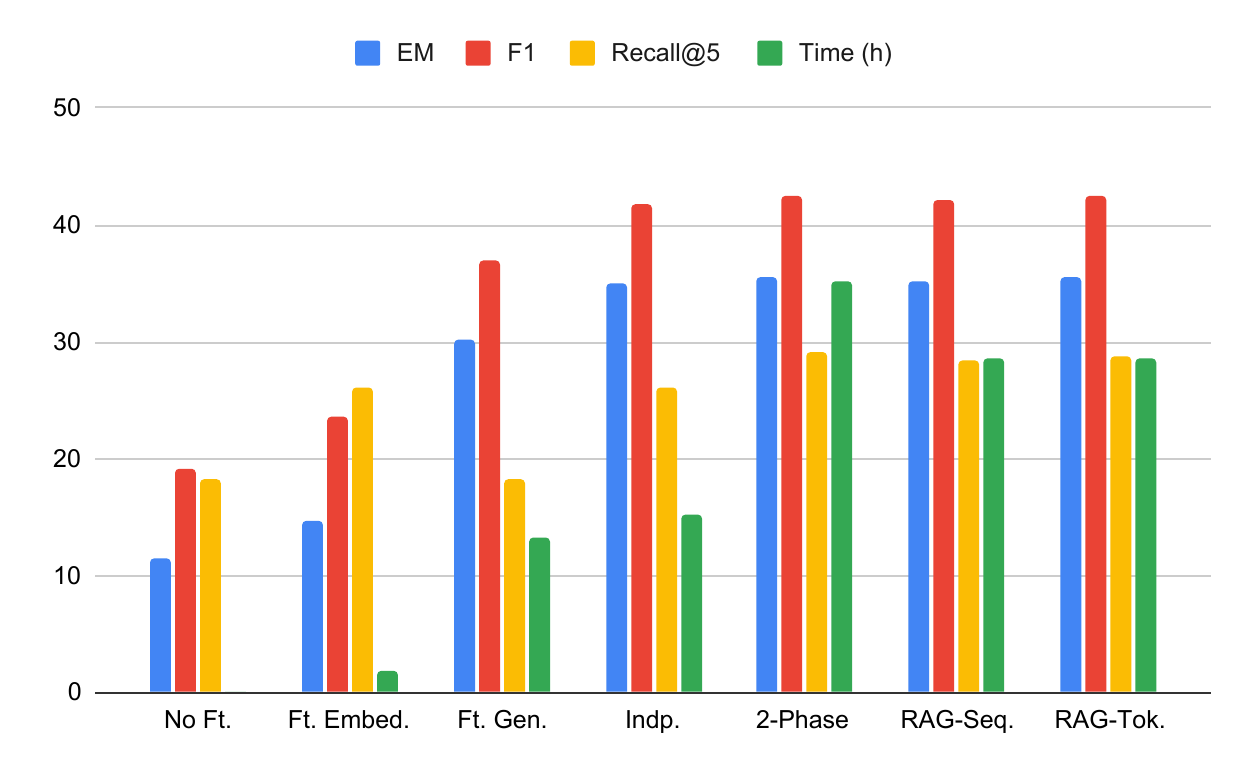}
  \caption{Validation performance metrics and time to fine-tune for different fine-tuning strategies, averaged across all four RAG pipelines and both HotPotQA and PopQA datasets.}
  \label{fig:results_chart}
\end{figure*}

\begin{figure*}[t]
    \centering
    \begin{tabular}{|c|cccc|cccc|}
        \hline
        \multirow{2}{*}{Method} & \multicolumn{4}{c|}{HotPotQA} & \multicolumn{4}{c|}{PopQA} \\
        & EM & F1 & Recall@5 & Time (h) & EM & F1 & Recall@5 & Time (h) \\
        \hline
        No Ft. & 10.3 & 19.8 & 19.1 & 0.0 & 12.6 & 18.6 & 17.4 & 0 \\
        Ft. Embed. & 11.1 & 20.8 & 21.4 & 3.5 & 18.2 & 26.6 & 30.8 & 0.4 \\
        Ft. Gen. & 28.4 & 39.4 & 19.1 & 23.8 & 32.1 & 34.7 & 17.4 & 2.9 \\
        Indp. & 29.3 & 40.2 & 21.4 & 27.4 & 40.6 & 43.2 & 30.8 & 3.2 \\
        2-Phase & 30.0 & 41.3 & 25.1 & 61.0 & 41.0 & 43.7 & 33.3 & 9.4 \\
        RAG-Seq. & 29.1 & 40.2 & 24.0 & 49.2 & 41.4 & 44.1 & 32.8 & 7.9 \\
        RAG-Tok. & 29.5 & 40.8 & 24.3 & 49.3 & 41.6 & 44.4 & 33.1 & 8.0 \\
        \hline
    \end{tabular}
    \caption{HotPotQA and PopQA validation performance metrics after fine-tuning and time to fine-tune for different fine-tuning strategies, averaged across all four RAG pipelines.}
    \label{table:results_table}
\end{figure*}

Here we evaluate and compare the performance of the RAG fine-tuning strategies described in the previous section for four RAG pipelines, each consisting of either an MPNet \citep{reimers-2019-sentence-bert} or MiniLM \citep{minilm} embedding model and either a LLaMA-3-8b-Instruct \citep{llama3modelcard} or Mistral-7b-Instruct-v0.1 \citep{jiang2023mistral7b} generator model. We fine-tune and evaluate on two datasets: HotPotQA \citep{yang-etal-2018-hotpotqa} and PopQA \citep{mallen2022not}. Our retrieval system uses the embedding model to retrieve the top $k=5$ most relevant documents from Wikipedia\footnote{https://huggingface.co/datasets/legacy-datasets/wikipedia}. We use the same chunking of Wikipedia as \citet{xiong2024benchmarking}, which contains $29.9$M chunks. We construct a vector database from the corpus using a FAISS index \citep{johnson2019billion}. Each experiment was conducted on a node with 8 NVIDIA A10 GPUs.

To minimize the computational expense of our experiments, in each experiment we fine-tune for only $1$ epoch (for the two-phase strategy, each model is fine-tuned for $1$ epoch) \citep{komatsuzaki2019one, egele2023one}. In all experiments, we use a linear learning rate schedule. To find near-optimal choices of learning rates, we perform a grid search over values between $10^{-8}$ and $10^{-4}$, with grid values separated roughly by factors of $3$: specifically, $10^{-8}$, $3 \times 10^{-8}$, $10^{-7}$, $3 \times 10^{-7}$, $10^{-6}$, $3 \times 10^{-6}$, $10^{-5}$, $3 \times 10^{-5}$, and $10^{-4}$. 

\subsection{Results}
The results of our experiments are in Table \ref{table:results_table} and illustrated in Figure \ref{fig:results_chart}. Each cell shows the validation exact match (EM), F1 metric, and Recall@5 for each experiment, averaged over the four RAG pipelines described at the beginning of this section. 
"No Ft." is the baseline RAG pipeline with no fine-tuning. "Ft. Embed." fine-tunes only the embedding model using context labels and the multiple negatives ranking loss. "Ft. Gen." fine-tunes only the generator model. "Indp." combines the independently fine-tuned embedding and generator models from "Ft. Embed." and "Ft. Gen." "2-Phase" is the two-phase fine-tuning strategy. "RAG-Seq." and "RAG-Tok." fine-tune the embedding and generator models jointly with RAG-Sequence and RAG-Token, respectively.

Comparing the "Baseline", "Ft. Embed.", and "Ft. Gen." experiments, we observe that fine-tuning the generator model alone significantly improves EM and F1 scores and that fine-tuning the embedding alone significantly improves Recall@5, with downstream benefits for EM and F1. We also observe that fine-tuning the generator model is much more computationally expensive than fine-tuning the embedding model using context labels. This is because the generator model is much larger than the embedding model, and so the latency of a single forward pass is much higher for the generator model than for the embedding model.

Comparing "Ft. Embed." to "2-Phase", "RAG-Seq.", and "RAG-Tok.", we observe that fine-tuning the embedding model using context labels may achieve worse Recall@5 compared to the end-to-end methods that do not use context labels. However, it may be possible to improve the results for our "Ft. Embed." experiment by using the cached variant of the multiple negatives ranking loss and increasing the batch size.


We observe that "Indp.", "2-Phase", "RAG-Sequence", and "RAG-Token" all achieve about the same EM and F1 scores. This suggests these strategies are about equally effective for fine-tuning a RAG pipeline. However, the strategies have significantly different computational cost: independent fine-tuning is the least expensive, followed by joint fine-tuning with RAG-Sequence or RAG-Token, followed by the two-phase fine-tuning strategy.

\section{Conclusion}
In this paper, we compared various strategies for fine-tuning the embedding and generator models of a RAG pipeline. From our experiments with four different RAG pipelines on HotPotQA and PopQA, we observed that independent, joint, and two-phase fine-tuning are all about equally effective for fine-tuning a RAG pipeline. While independent fine-tuning is computationally less expensive, joint fine-tuning and two-phase fine-tuning have the benefit of not requiring context labels to perform fine-tuning. In addition, two-phase fine-tuning allows for a more efficient hyperparameter search for the embedding and generator model learning rates compared to joint fine-tuning.
\clearpage

\section*{Limitations}
In order to maximize the end-to-end performance of each fine-tuning strategy, we used a grid search to find near-optimal choices of the learning rates for the embedding and generator models. However, it may be possible to further increase end-to-end performance by additionally performing  hyperparameter optimizations over the number of training epochs and the training batch size. In particular, it may be possible to improve the end-to-end performance achieved in the "Ft. Embed." experiments, which fine-tune the embedding model by optimizing the multiple negatives ranking loss, by increasing the training batch size to a number much larger than 8. 

We perform our fine-tuning experiments using a basic RAG pipeline setup. However, more complex RAG pipelines are common in practice, e.g., pipelines that perform context document re-ranking after the document retrieval step, or pipelines that perform multiple document retrieval steps to answer multi-hop questions. It remains unclear how introducing these complexities to the RAG pipeline might impact the effectiveness of each of the fine-tuning strategies discussed in this paper.

\bibliography{custom}

\clearpage
\appendix

\section{Prompt}
\label{sec:prompt}
In all experiments, we use the following prompt for the generative model to generate an answer given a question and concatenated context documents.
\begin{verbatim}
prompt = """You are a helpful general \
knowledge expert. Answer the following \
question using the relevant context. Use \
as few words as possible.

### Context:
{context}

### Question:
{question}

### Answer:
"""
\end{verbatim}

\begin{figure*}[ht]
    \centering
    \includegraphics[width=0.8\textwidth]{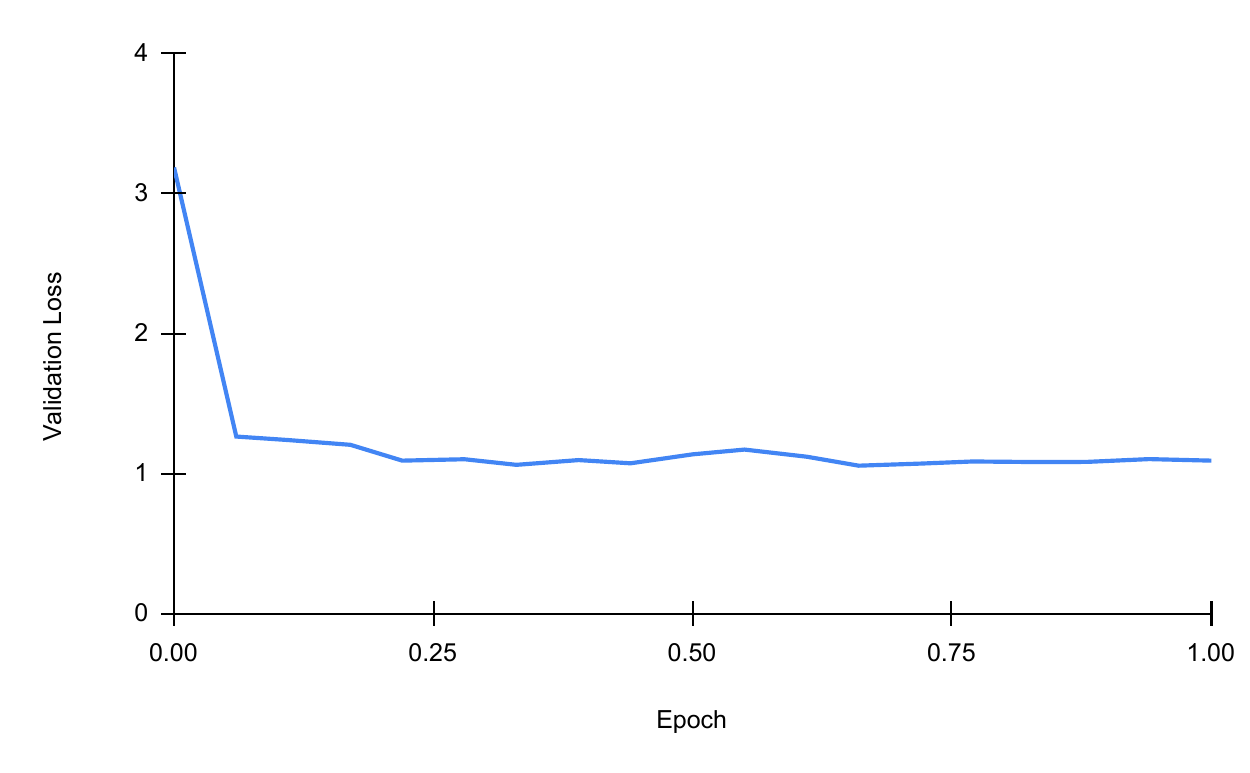}
    \caption{Validation loss convergence plot for fine-tuning a RAG pipeline consisting of a MiniLM embedding model and LLaMA-3-8b generator model on HotPotQA with joint fine-tuning. The validation loss converges quickly during fine-tuning, well within the 1 epoch fine-tuning period.}
    \label{fig:validation_loss_v_epochs}
\end{figure*}


\begin{figure*}[ht]
  \centering
    \begin{tabular}{|cc|}
        \hline
        Model Name & \# Params \\
        \hline
        MiniLM & 22.7M \\
        MPNet & 109M \\
        Mistral-7b & 7.24B \\
        LLaMA3-8b & 8.03B \\
        \hline
    \end{tabular}
  \caption{Number of parameters in each model used in this paper.}
  \label{fig:model_params}
\end{figure*}

\begin{figure*}[ht]
    \centering
    \begin{tabular}{|c|c|c|cccccc|}
        \hline
        Embed. & Gen. & \multirow{2}{*}{Method} & \multicolumn{6}{c|}{HotPotQA} \\
        Model & Model & & EM & F1 & Recall@5 & Time(h) & Embed. LR & Gen. LR \\
        \hline
\multirow{7}{*}{MiniLM} & \multirow{7}{*}{LLaMA3-8b} & No Ft. & 15.3 & 24.6 & 19.5 & 0.0 & N/A & N/A \\
& & Ft. Embed & 16.5 & 26.0 & 21.3 & 1.5 & 1E-06 & N/A \\
& & Ft. Gen & 29.9 & 41.2 & 19.5 & 21.8 & N/A & 1E-05 \\
& & Indp. & 30.5 & 41.7 & 21.3 & 23.3 & 1E-06 & 1E-05 \\
& & 2-Phase & 30.8 & 42.4 & 23.7 & 35.2 & 3E-08 & 1E-05 \\
& & RAG-Seq. & 27.8 & 38.5 & 22.9 & 45.9 & 3E-08 & 1E-05 \\
& & RAG-Tok. & 30.0 & 41.4 & 23.2 & 46.0 & 3E-08 & 1E-05 \\
\hline
\multirow{7}{*}{MiniLM} & \multirow{7}{*}{Mistral-7b} & No Ft. & 5.5 & 15.2 & 19.5 & 0.0 & N/A & N/A \\
& & Ft. Embed & 6.2 & 15.7 & 21.3 & 1.5 & 1E-06 & N/A \\
& & Ft. Gen & 26.8 & 37.5 & 19.5 & 24.6 & N/A & 1E-05 \\
& & Indp. & 27.9 & 38.5 & 21.3 & 26.1 & 1E-06 & 1E-05 \\
& & 2-Phase & 27.7 & 38.7 & 23.0 & 36.6 & 3E-08 & 1E-05 \\
& & RAG-Seq. & 27.5 & 38.4 & 22.6 & 49.9 & 3E-08 & 1E-05 \\
& & RAG-Tok. & 26.8 & 37.3 & 22.3 & 49.8 & 3E-08 & 1E-05 \\
\hline
\multirow{7}{*}{MPNet} & \multirow{7}{*}{LLaMA3-8b} & No Ft. & 15.1 & 24.5 & 18.6 & 0.0 & N/A & N/A \\
& & Ft. Embed & 16.0 & 25.9 & 21.5 & 5.5 & 1E-06 & N/A \\
& & Ft. Gen & 29.8 & 41.0 & 18.6 & 22.9 & N/A & 3E-06 \\
& & Indp. & 30.7 & 41.8 & 21.5 & 28.4 & 1E-06 & 3E-06 \\
& & 2-Phase & 32.1 & 43.8 & 27.3 & 37.9 & 3E-08 & 3E-06 \\
& & RAG-Seq. & 31.8 & 43.7 & 25.7 & 48.7 & 3E-08 & 3E-06 \\
& & RAG-Tok. & 31.9 & 44.0 & 26.4 & 49.1 & 3E-08 & 3E-06 \\
\hline
\multirow{7}{*}{MPNet} & \multirow{7}{*}{Mistral-7b} & No Ft. & 5.4 & 15.0 & 18.6 & 0.0 & N/A & N/A \\
& & Ft. Embed & 5.7 & 15.6 & 21.5 & 5.5 & 1E-06 & N/A \\
& & Ft. Gen & 27.2 & 37.8 & 18.6 & 26.0 & N/A & 1E-05 \\
& & Indp. & 28.1 & 38.8 & 21.5 & 31.6 & 1E-06 & 1E-05 \\
& & 2-Phase & 29.4 & 40.6 & 26.4 & 39.1 & 3E-08 & 1E-05 \\
& & RAG-Seq. & 29.1 & 40.4 & 24.7 & 52.4 & 3E-08 & 1E-05 \\
& & RAG-Tok. & 29.2 & 40.3 & 25.3 & 52.5 & 3E-08 & 1E-05 \\
        \hline
    \end{tabular}
    \caption{HotPotQA validation performance metrics after fine-tuning, time to fine-tune, and learning rates used for different fine-tuning strategies and RAG pipelines.}
    \label{table:results_table_hotpotqa}
\end{figure*}

\begin{figure*}[ht]
    \centering
    \begin{tabular}{|c|c|c|cccccc|}
        \hline
        Embed. & Gen. & \multirow{2}{*}{Method} & \multicolumn{6}{c|}{PopQA} \\
        Model & Model & & EM & F1 & Recall@5 & Time(h) & Embed. LR & Gen. LR \\
        \hline
\multirow{7}{*}{MiniLM} & \multirow{7}{*}{LLaMA3-8b} & No Ft. & 17.3 & 23.4 & 17.9 & 0.0 & N/A & N/A \\
& & Ft. Embed & 23.6 & 31.1 & 28.5 & 0.1 & 1E-05 & N/A \\
& & Ft. Gen & 34.6 & 37.4 & 17.9 & 2.5 & N/A & 1E-05 \\
& & Indp. & 40.8 & 43.7 & 28.5 & 2.6 & 1E-05 & 1E-05 \\
& & 2-Phase & 41.1 & 44.0 & 30.7 & 6.3 & 3E-07 & 1E-05 \\
& & RAG-Seq. & 40.6 & 43.6 & 30.1 & 7.2 & 3E-07 & 1E-05 \\
& & RAG-Tok. & 41.8 & 44.3 & 30.9 & 7.3 & 3E-07 & 1E-05 \\
\hline
\multirow{7}{*}{MiniLM} & \multirow{7}{*}{Mistral-7b} & No Ft. & 8.9 & 15.3 & 17.9 & 0.0 & N/A & N/A \\
& & Ft. Embed & 12.1 & 20.4 & 28.5 & 0.1 & 1E-05 & N/A \\
& & Ft. Gen & 30.9 & 33.4 & 17.9 & 2.7 & N/A & 3E-05 \\
& & Indp. & 37.5 & 40.5 & 28.5 & 2.8 & 1E-05 & 3E-05 \\
& & 2-Phase & 38.6 & 41.5 & 31.3 & 6.5 & 3E-08 & 3E-05 \\
& & RAG-Seq. & 39.5 & 42.3 & 30.6 & 7.7 & 3E-08 & 3E-05 \\
& & RAG-Tok. & 39.9 & 42.4 & 31.4 & 7.8 & 3E-08 & 3E-05 \\
\hline
\multirow{7}{*}{MPNet} & \multirow{7}{*}{LLaMA3-8b} & No Ft. & 16.0 & 21.6 & 16.9 & 0.0 & N/A & N/A \\
& & Ft. Embed & 25.1 & 33.4 & 33.1 & 0.6 & 3E-05 & N/A \\
& & Ft. Gen & 33.6 & 36.1 & 16.9 & 3.0 & N/A & 1E-04 \\
& & Indp. & 43.0 & 45.5 & 33.1 & 3.5 & 3E-05 & 1E-04 \\
& & 2-Phase & 43.2 & 45.9 & 35.8 & 6.4 & 3E-07 & 1E-04 \\
& & RAG-Seq. & 44.0 & 46.5 & 35.4 & 8.1 & 3E-07 & 1E-04 \\
& & RAG-Tok. & 42.4 & 46.1 & 35.2 & 8.1 & 3E-07 & 1E-04 \\
\hline
\multirow{7}{*}{MPNet} & \multirow{7}{*}{Mistral-7b} & No Ft. & 8.2 & 14.2 & 16.9 & 0.0 & N/A & N/A \\
& & Ft. Embed & 12.0 & 21.2 & 33.1 & 0.6 & 3E-05 & N/A \\
& & Ft. Gen & 29.2 & 31.9 & 16.9 & 3.3 & N/A & 3E-05 \\
& & Indp. & 40.8 & 43.2 & 33.1 & 3.9 & 3E-05 & 3E-05 \\
& & 2-Phase & 40.9 & 43.5 & 35.5 & 6.9 & 3E-07 & 3E-05 \\
& & RAG-Seq. & 41.5 & 44.1 & 35.2 & 8.7 & 3E-07 & 3E-05 \\
& & RAG-Tok. & 42.2 & 44.9 & 35.0 & 8.6 & 3E-07 & 3E-05 \\
        \hline
    \end{tabular}
    \caption{PopQA validation performance metrics after fine-tuning, time to fine-tune, and learning rates used for different fine-tuning strategies and RAG pipelines.}
    \label{table:results_table_popqa}
\end{figure*}


\end{document}